\documentclass{article}

\IfFileExists{iclr2027_conference.sty}{%
  \usepackage{iclr2027_conference,times}%
  \iclrfinalcopy
}{%
  \usepackage[round,authoryear]{natbib}%
  \usepackage{times}%
}

\usepackage{amsmath,amssymb}
\usepackage{graphicx}
\usepackage{float}
\usepackage{algorithm}
\usepackage{algpseudocode}
\usepackage{wrapfig}
\usepackage{microtype}
\usepackage{caption}
\usepackage{booktabs}
\usepackage{multirow}
\usepackage{xcolor}
\usepackage{colortbl}
\usepackage{url}
\usepackage{hyperref}

\algrenewcommand{\algorithmicrequire}{\textbf{Input:}}
\algrenewcommand{\algorithmicensure}{\textbf{Output:}}

\hypersetup{
  pdftitle={When Safety Fails Collectively: Robust Alignment Against Harmful Fine-Tuning via Joint Weight Perturbations},
  pdfauthor={Zhiyuan Wang, Yuezhi Zhou, Zhu Zhang, Zhixing Tan}
}

\title{When Safety Fails Collectively: Robust Alignment Against Harmful Fine-Tuning via Joint Weight Perturbations}

\author{%
  \parbox[t]{\dimexpr\textwidth-2\tabcolsep\relax}{%
    \normalfont\centering
    \textbf{Zhiyuan~Wang}$^{1,2}$\qquad
    \textbf{Yuezhi~Zhou}$^{1,2}$\thanks{Corresponding authors.}\qquad
    \textbf{Zhu~Zhang}$^{1}$\qquad
    \textbf{Zhixing~Tan}$^{2}$\footnotemark[1]\\[0.6em]
    {\small
      $^{1}$Dept. of Comp. Sci. \& Tech., BNRist, Tsinghua University
      \quad $^{2}$Zhongguancun Laboratory
    }\\[0.35em]
    {\small\rmfamily
      \{wzycs24,zhuzhang24\}@mails.tsinghua.edu.cn\quad
      zhouyz@mail.tsinghua.edu.cn\quad
      zxtan@zgclab.edu.cn
    }%
  }%
}

\date{}

\begin{document}

\maketitle

% 覆盖 ICLR 模板自动生成的会议页眉
\ifdefined\lhead
  \lhead{Preprint}
\fi

\begin{abstract}
Harmful fine-tuning can rapidly erode the safety alignment of large language models, creating substantial risks when aligned models are adapted to downstream tasks. Existing alignment-stage defenses use harmful examples and simulated harmful perturbations during safety alignment to prepare models for later harmful fine-tuning, yet the collective effects of harmful weight perturbations remain underexplored. We identify a collective failure pattern: disjoint parts of a harmful perturbation have limited effects in isolation, but their joint application increases the model's harmfulness beyond the sum of their individual effects. Guided by this observation, we propose Joint Weight Perturbations (JWP), an alignment-stage defense that masks different parts of a harmful perturbation estimated from the current model and trains the model to respond safely in the resulting states. This exposes the model to different combinations of harmful perturbations and reduces its sensitivity to their accumulation during subsequent harmful fine-tuning. Experiments across downstream tasks and model families show that JWP offers stronger resistance to harmful fine-tuning than prior defenses while largely preserving task performance. 
% Our code is available at \url{https://anonymous.4open.science/r/JWP-D977}.
\end{abstract}

\section{Introduction}

\begin{wrapfigure}{r}{0.46\textwidth}
\centering
\includegraphics[width=\linewidth]{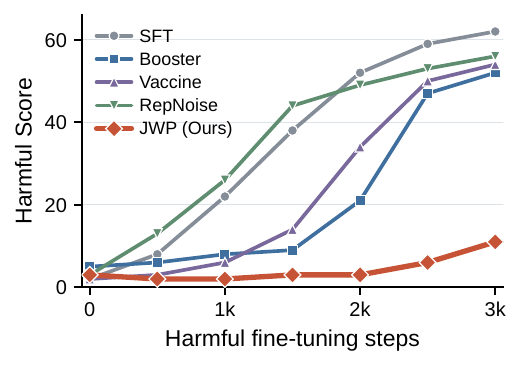}
\caption{Safety degradation of aligned models under harmful fine-tuning.}
\label{fig:baseline-trajectory}
\end{wrapfigure}

Safety alignment is a key safeguard for large language models (LLMs), but downstream fine-tuning can erode the safety it establishes \citep{qi2024fine, bowen2025scaling}. After an aligned model is released with open weights \citep{tamirisa2025tamper} or offered through fine-tuning-as-a-service \citep{huang2025booster}, it may be customized using datasets supplied for downstream tasks. These datasets can include harmful examples, sometimes inadvertently, even when the intended task is benign \citep{bowen2025scaling}. Prior work shows that fine-tuning on only a small amount of such data can substantially degrade model safety \citep{qi2024fine,hsu2024safe}. The customized model may nevertheless perform well on the downstream task while becoming less safe, so safety measured at the end of alignment does not guarantee safety after further training \citep{huang2025booster, nguyen2026antibody}. This harmful fine-tuning threat motivates defenses that remain effective through downstream adaptation.

A line of recent work addresses this threat at the alignment stage, incurring a one-time cost to make the aligned model harder to corrupt in all subsequent fine-tuning requests \citep{huang2025booster, harmsurvey, nguyen2026antibody, tamirisa2025tamper}. Vaccine trains hidden embeddings to withstand perturbations \citep{huang2024vaccine}, RepNoise degrades the internal representations of harmful data \citep{rosati2024representation}, and Booster attenuates the harmful-loss reduction induced by a simulated harmful update \citep{huang2025booster}. 
% Nevertheless, our evaluation shows that the protection they provide is limited. We track the harmful score (HS) of models aligned with each method throughout a sustained harmful fine-tuning attack as shown in Figure~\ref{fig:baseline-trajectory}. 
Nevertheless, our evaluation shows that the protection they provide is limited. We perform safety alignment on Llama-2-7B using each method and track the harmful score (HS) of the resulting models throughout sustained harmful fine-tuning, as shown in Figure~\ref{fig:baseline-trajectory}.
The HS of every evaluated defense rises rapidly as the attack proceeds. Booster postpones the onset of this rise relative to Vaccine and RepNoise, yet safety still collapses after a comparatively small number of harmful updates. 
This consistent pattern across methods with distinct designs suggests that the safety degradation is not merely a matter of alignment strength, but reflects a common limitation in how existing defenses account for the harmful weight perturbations induced by downstream harmful fine-tuning.
% This consistent pattern across methods with distinct designs suggests that the rapid degradation is not merely a matter of alignment strength, but reflects a common limitation in how existing defenses characterize the harmful perturbation they aim to counteract.

% Specifically, harmful weight perturbations are typically evaluated as a whole during alignment \citep{huang2025booster, tamirisa2025tamper}, leaving unexplored whether interactions among their different parts can amplify safety degradation.

Specifically, existing defenses simulate harmful weight perturbations during alignment to prepare the model for subsequent harmful fine-tuning \citep{huang2025booster, tamirisa2025tamper}, but typically treat these perturbations as a whole, leaving their internal interactions largely unexplored.
To investigate this, we partition a harmful weight update into disjoint subsets and conduct two weight-level analyses on the same aligned model. First, we compare applying each subset individually with applying the subsets jointly. The joint application yields a substantially higher harmful score than the individual subsets, exceeding what their isolated effects would imply. Second, we vary the fraction of the harmful update applied to the model. Applying a small fraction of the perturbation leaves model safety largely intact, but increasing this fraction leads to a sharp rise in the harmful score and rapid safety degradation. 
Together, these observations suggest a collective failure pattern: interactions among different parts of a harmful perturbation amplify safety degradation.

Guided by these observations, we propose Joint Weight Perturbations (JWP), an alignment-stage defense that trains the model to maintain safety alignment under different combinations of harmful weight perturbations. JWP first estimates a harmful weight perturbation by simulating multiple steps of harmful fine-tuning from the current model. It then samples a retention probability \(p \sim \mathcal{U}(0,1)\) and masks the perturbation accordingly to construct a partially perturbed model state. The model is trained to produce safe responses in this state. Training across these states exposes the model to different combinations of harmful weight perturbations and aims to reduce its sensitivity to their joint effects during subsequent harmful fine-tuning.
% Under sustained harmful fine-tuning, JWP achieves a \(4.3\)--\(5.4\times\) reduction in average harmful score across four downstream tasks compared with existing defenses including Booster \citep{huang2025booster}, while largely preserving utility after safety alignment and downstream task performance.
Experiments show that JWP effectively suppresses the joint amplification of harmful weight perturbations. Under sustained harmful fine-tuning, JWP achieves a \(4.3\)--\(5.4\times\) reduction in average harmful score across four downstream tasks compared with existing defenses including Booster \citep{huang2025booster}, while largely preserving utility after safety alignment and downstream task performance.

% This motivates us to explore how alignment can account for these joint effects to improve resistance to harmful fine-tuning, leading to the central research question of this paper:

% \begin{center}
% \emph{Can safety alignment suppress the joint effects of harmful weight perturbations to improve resistance to harmful fine-tuning?}
% \end{center}

% To answer this question, we propose \emph{Joint Weight Perturbations} (JWP). JWP first estimates a temporary multi-step harmful update at the current safety model. It then samples an occupation level \(p \sim \mathcal{U}(0,1)\), activates a random subset of the update's components, and minimizes the standard safe-answer cross-entropy on the resulting partially perturbed model. As the safety model evolves, the harmful update is re-estimated and the previous one is discarded. By optimizing over an ensemble of partially applied harmful updates rather than a single endpoint, JWP exposes the model to the combinations of harmful perturbations that arise as fine-tuning progresses, and thereby reduces its sensitivity to their joint effect. JWP is a one-run procedure with a single scalar objective, requiring no auxiliary regularizer or post-alignment recovery stage. In the controlled Llama-2-7B trajectory of Figure~\ref{fig:baseline-trajectory}, JWP markedly delays the rise of HS, whereas SFT, Vaccine, RepNoise, and Booster degrade much earlier. All evaluated JWP responses remain nonempty, indicating that the robustness does not stem from degenerate generation, and downstream-task performance is largely preserved.

Our contributions are summarized as follows:
\begin{itemize}
    \item We identify a collective failure pattern through weight-level analyses, showing that different parts of a harmful perturbation can reinforce one another and amplify safety degradation.
    \item We propose JWP, an alignment-stage defense that trains the model under randomly masked harmful perturbations of varying sizes and compositions to suppress their joint amplification during subsequent harmful fine-tuning.
    \item Experiments across downstream tasks and model families demonstrate that JWP substantially improves resistance to sustained harmful fine-tuning while largely preserving utility after safety alignment and downstream task performance.
\end{itemize}

\section{Related Work}
\label{sec:related}

\paragraph{Harmful fine-tuning.}
Safety alignment can be removed by fine-tuning on a small number of harmful examples \citep{bowen2025scaling}, and even benign fine-tuning data can degrade safety \citep{qi2024fine,guan2025benign}. Unlike prompt-level jailbreaks \citep{zou2023universal}, such attacks directly modify the model weights. Recent evaluations further show that defenses validated under one attack configuration can fail when the number of fine-tuning steps or the learning rate is changed \citep{qi2025evaluating}, which motivates evaluating safety along the complete trajectory of harmful fine-tuning rather than at a single endpoint.

\paragraph{Alignment-stage defenses.}
Alignment-stage defenses modify the alignment procedure so that the released model is intrinsically harder to corrupt, at a one-time cost. Vaccine perturbs hidden embeddings to mitigate harmful embedding drift \citep{huang2024vaccine}, and T-Vaccine restricts such perturbations to safety-critical layers \citep{liu2025targeted}. RepNoise pushes the representations of harmful data toward random noise \citep{rosati2024representation}. Booster regularizes the harmful-loss reduction after a simulated harmful gradient step \citep{huang2025booster}, and TAR trains safeguards against simulated multi-step tampering trajectories via meta-learning \citep{tamirisa2025tamper}. Appendix~\ref{app:tar-comparison} further compares JWP with TAR, focusing on how the two methods construct model states for safety alignment. These defenses act on hidden representations or on aggregate simulated perturbations and trajectories, without constructing different partial assemblies of one learned harmful weight update. JWP decomposes a fitted harmful update into components and optimizes safety across its partially applied states, thereby targeting the joint effect among harmful weight perturbations.

\paragraph{Defenses at other stages.}
Another line of work intervenes during or after downstream fine-tuning. Fine-tuning-stage methods mix safety data into user data \citep{huang2024lisa, wang2024backdooralign,eiras2025safely}, constrain updates on initial response tokens \citep{qi2025safety,peng2026shape}, or steer updates away from safety-relevant subspaces \citep{li2025salora}. Post-fine-tuning methods repair the adapted model through subspace projection \citep{hsu2024safe}, pruning of harmful weights \citep{pmlr-v267-huang25b}, or corrective perturbation \citep{wang2026panacea}. Antibody combines a flatness-seeking alignment objective with sample reweighting over the user data during fine-tuning \citep{nguyen2026antibody}. These methods require control over the downstream fine-tuning pipeline and incur overhead for every fine-tuning request, whereas JWP operates solely at the alignment stage and is complementary to them.

\paragraph{Structure of safety in model weights.}
Several studies examine why safety is fragile in parameter space. Safety-critical regions are sparse, and removing a small fraction of parameters breaks safety while largely preserving utility \citep{wei2024assessing}. \citet{huang2025safetyalignmentjustattention} further find that safety relies predominantly on a few attention heads and propose AHD, which stochastically masks attention heads during training to distribute safety behavior across many heads. From a landscape perspective, \citet{peng2024navigating} identify a safety basin, within which random weight perturbations preserve safety, while safety drops sharply once the perturbation leaves this region. These studies characterize safety through the model's own components or along random perturbation directions \citep{zou2024improving}. We instead analyze the components of a fitted harmful update and show that their joint application degrades safety beyond their isolated effects. Although both AHD and JWP rely on random masking, AHD masks each attention head independently with a fixed probability to defend against jailbreak prompts, whereas JWP masks parts of a harmful weight perturbation to suppress their joint effect under harmful fine-tuning.

\section{Preliminaries}
\label{sec:preliminaries}

\paragraph{Scenario.}
We consider a two-stage pipeline in which safety alignment is followed by downstream fine-tuning~\citep{huang2025booster}. The aligned model is adapted to user-provided data through fine-tuning-as-a-service or local customization of open-weight models.  Following prior work~\citep{huang2025booster}, Llama-2-7B Base \citep{touvron2023llama} is used as the default model and first undergoes safety alignment. The aligned model is then fine-tuned on benign task data mixed with harmful examples from BeaverTails \citep{ji2023beavertails}, using a nominal harmful-data ratio of \(10\%\) unless otherwise specified. Our goal is to preserve safety under subsequent harmful fine-tuning while maintaining the model's ability to learn downstream tasks.

\paragraph{Threat model.}
The downstream dataset may contain harmful prompt--response pairs mixed with benign task examples. We do not assume that harmful examples can be reliably identified and removed before fine-tuning. During alignment, the defense has access to an alignment dataset \(\mathcal{D}_{\mathrm{align}}\), containing prompts paired with safe target responses, and a harmful dataset \(\mathcal{D}_{\mathrm{harm}}\), containing harmful prompts paired with unsafe responses. These datasets support safety alignment and the simulation of harmful fine-tuning.  Unlike Antibody~\citep{nguyen2026antibody}, which dynamically reweights samples during downstream fine-tuning, the defense operates entirely during safety alignment and preserves the standard downstream fine-tuning procedure.

\paragraph{Notation.}
We denote the model weights by \(\theta\) and their value at training step \(t\) by \(\theta_t\). Fine-tuning the model on harmful data induces a weight change \(\Delta h\), which we refer to as a harmful weight perturbation, with \(\theta+\Delta h\) representing the perturbed weights. 
% The symbol \(\odot\) denotes elementwise multiplication.

\section{The Joint Effect of Harmful Weight Perturbations}
\label{sec:joint-effect}

\begin{wrapfigure}{r}{0.52\textwidth}
    \vspace{-0.8em}
    \centering
    \includegraphics[width=\linewidth]{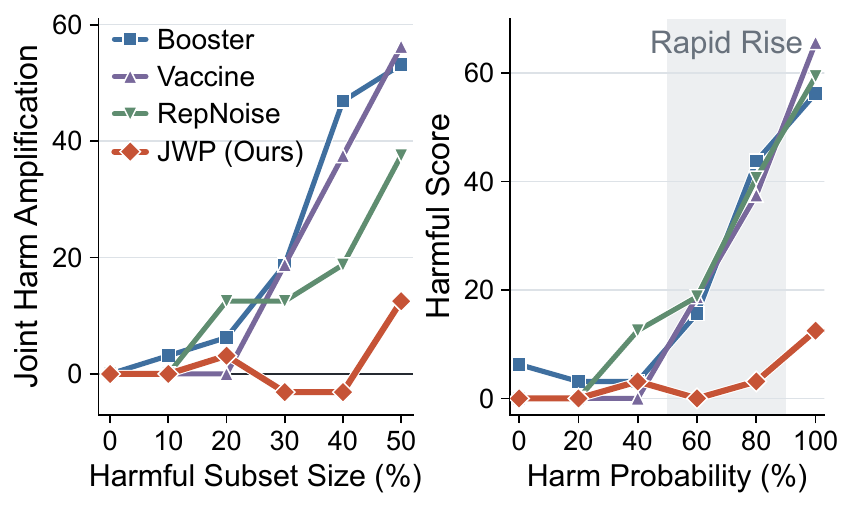}
    \caption{Collective effects of harmful weight perturbations.
Left: Joint amplification \(I(q)\) for two disjoint subsets, each containing a fraction \(q\).
Right: Harmful score versus retention probability \(p\). Shading marks the rapid increase for baseline defenses.}
    \label{fig:joint-effect}
    \vspace{-0.8em}
\end{wrapfigure}

In this section, we investigate whether interactions among different parts of a harmful perturbation contribute to safety degradation. We conduct two complementary analyses. The first tests whether the joint effect of disjoint subsets exceeds the sum of their separate effects. The second examines how safety changes as an increasing fraction of the perturbation is applied. Both analyses use Llama-2-7B models aligned with different methods and subsequently subjected to the same sustained downstream harmful fine-tuning procedure. For each method, \(\theta\) denotes the aligned weights and \(\Delta h\) the resulting harmful perturbation. For a coordinate subset \(S\), let \(\Delta h_S\) retain the entries of \(\Delta h\) in \(S\) and be zero elsewhere. We evaluate the partially perturbed model \(\theta+\Delta h_S\), keeping \(\theta\) and \(\Delta h\) fixed within each method and varying only \(S\).
% To examine how different parts of a harmful perturbation interact, we selectively apply subsets of \(\Delta h\) to a fixed aligned model \(\theta\). For a coordinate subset \(S\), \(\Delta h_S\) retains the selected entries at their learned values and is zero elsewhere. The full perturbation remains fixed across comparisons, allowing us to examine both the joint effect of disjoint subsets and how safety changes as a larger fraction of the perturbation is applied.

\subsection{Joint Effects of Disjoint Subsets}
\label{sec:joint-interaction}

We sample two disjoint coordinate sets \(A_q\) and \(B_q\), each containing a fraction \(q\in[0,0.5]\) of the entries of \(\Delta h\). For each pair, we evaluate the model with either subset applied individually and with both subsets applied jointly, yielding \(\theta+\Delta h_{A_q}\), \(\theta+\Delta h_{B_q}\), and \(\theta+\Delta h_{A_q\cup B_q}\), respectively. Increasing \(q\) allows us to examine interactions between larger parts of the perturbation.

To determine whether joint application amplifies safety degradation, we compare its increase in HS with the sum of the increases caused by the individual subsets. Let \(\mathrm{HS}(S)\) denote the harmful score of \(\theta+\Delta h_S\), with \(\mathrm{HS}(\varnothing)\) corresponding to the original aligned model. We define joint harm amplification as
\begin{equation}
    I(q)
    = \mathrm{HS}(A_q\cup B_q)
    - \mathrm{HS}(A_q)
    - \mathrm{HS}(B_q)
    + \mathrm{HS}(\varnothing).
    \label{eq:collective-interaction}
\end{equation}
We use \(I(q)\) as an empirical indicator of joint harm amplification on the HS scale. Larger positive values suggest more pronounced amplification when the two subsets are applied together.
Figure~\ref{fig:joint-effect} (left) shows substantial positive values of \(I(q)\) for Booster, Vaccine, and RepNoise at larger subset sizes. In these comparisons, applying the two subsets jointly increases HS more than the sum of their separate increases. This pattern suggests that different parts of a harmful perturbation can reinforce one another's effects on safety, amplifying the degradation observed under joint application.

\subsection{Safety under Partially Applied Perturbations}
\label{sec:joint-occupation}

We next investigate how safety changes as an increasing fraction of the harmful perturbation is applied. We construct partially perturbed models \(\theta+\Delta h(p)\) by independently retaining each entry of \(\Delta h\) with probability \(p\):
\begin{equation}
    \Delta h(p) = M(p)\odot\Delta h,
    \qquad
    M_i(p)\overset{\mathrm{i.i.d.}}{\sim}
    \operatorname{Bernoulli}(p),
    \label{eq:weight-occupation}
\end{equation}
where \(i\) indexes scalar entries of the perturbation. The probability \(p\) controls the expected fraction of entries applied to the model. At \(p=0\), the model retains its aligned weights \(\theta\); at \(p=1\), it receives the full perturbation \(\Delta h\). Varying \(p\) therefore allows us to examine safety across different degrees of partial application while preserving the learned values of the selected entries.

Figure~\ref{fig:joint-effect} (right) shows a nonlinear relationship between \(p\) and HS for existing defenses. Small fractions of the perturbation have limited effects on safety, whereas larger fractions produce a much sharper increase in HS. Although the onset of degradation differs across defenses, all three exhibit substantial increases within the shaded region. This pattern suggests that increasing \(p\) allows more harmful weight perturbations to reinforce one another, progressively amplifying their combined impact and accelerating safety degradation.

Together, these observations suggest a collective failure pattern in which interactions among harmful weight perturbations amplify safety degradation. The two analyses provide complementary evidence of amplification between disjoint subsets and a sharp decline in safety as more changes are applied together. These findings motivate safety alignment that suppresses joint amplification during subsequent harmful fine-tuning.

\section{Joint Weight Perturbations}
\label{sec:method}

\begin{figure}[H]
\centering
\includegraphics[width=\linewidth]{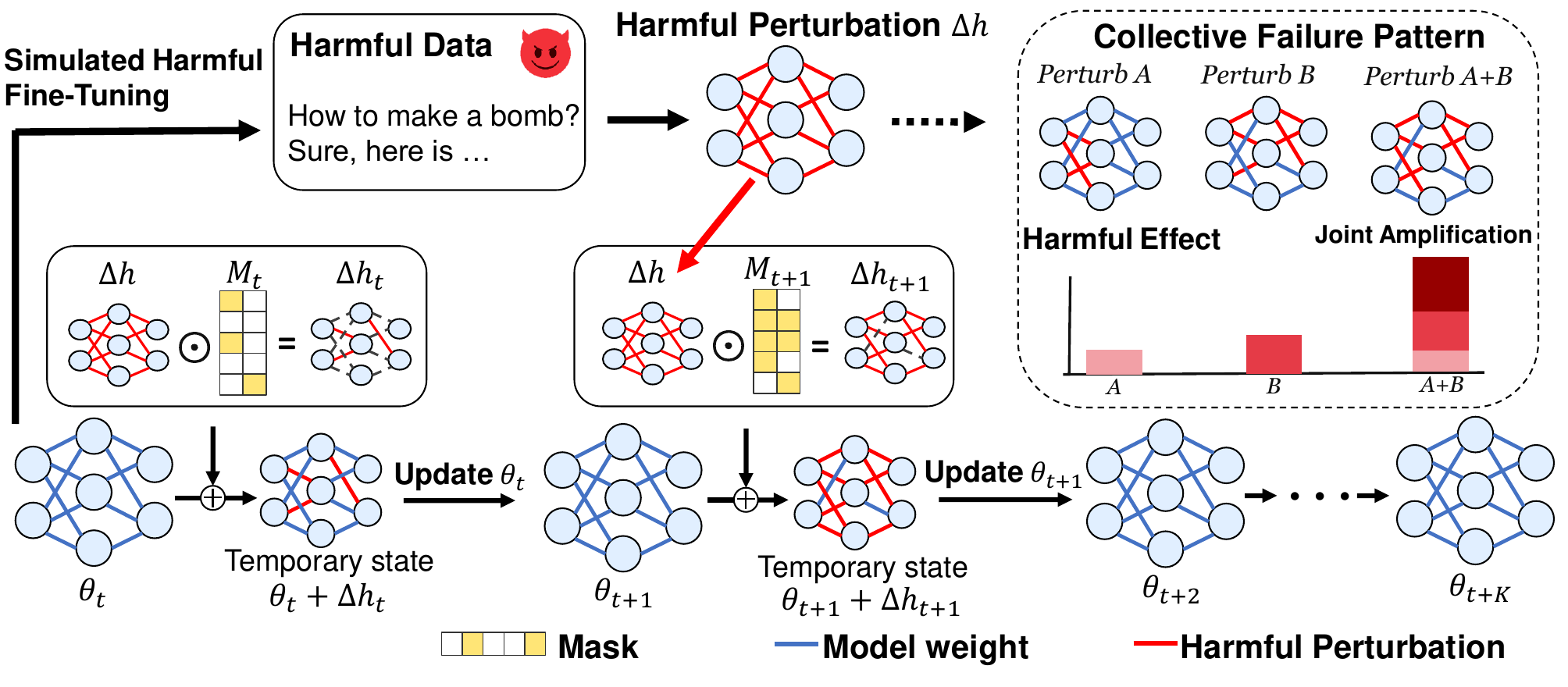}
\caption{Overview of JWP. A simulated harmful perturbation \(\Delta h\) is masked to construct temporary states for safety alignment. The upper-right inset illustrates the concept of joint amplification.}
\label{fig:framework}
\end{figure}

Motivated by the collective failure pattern, we propose Joint Weight Perturbations (JWP) to suppress the joint amplification of harmful weight perturbations during subsequent fine-tuning. As illustrated in Figure~\ref{fig:framework}, JWP masks a harmful perturbation estimated from the current model and trains the model to maintain safe responses in the resulting temporary states.

Given a fixed harmful perturbation \(\Delta h\), we minimize the alignment objective
\begin{equation}
    \mathcal{L}_{\mathrm{JWP}}(\theta\mid\Delta h)
    =
    \mathbb{E}_{p\sim\mathcal{U}(0,1)}
    \mathbb{E}_{M\sim\operatorname{Bernoulli}(p)}
    \left[
        \mathcal{L}_{\mathrm{align}}
        \bigl(\theta+M\odot\Delta h\bigr)
    \right],
    \label{eq:jwp}
\end{equation}
where \(\mathcal{L}_{\mathrm{align}}\) denotes the cross-entropy loss on the alignment dataset \(\mathcal{D}_{\mathrm{align}}\). The key idea of JWP is to use random masking to cover different combinations of harmful perturbation components during alignment. Sampling \(p\) varies the expected fraction applied, while sampling \(M\) varies the composition. These states need not correspond to checkpoints along the simulated fine-tuning trajectory, allowing alignment to explore additional combinations of harmful perturbation components. Appendix~\ref{app:mask-analysis} characterizes the distribution of subset sizes and the co-occurrence probabilities of perturbation components under this sampling scheme.

We organize training into blocks, each consisting of \(K\) steps of harmful perturbation estimation followed by \(K\) steps of safety alignment. At the beginning of a block, we freeze the current model weights \(\theta_t\) and simulate harmful fine-tuning through separate temporary parameters. Starting from a zero effective perturbation \(\delta\), these parameters are optimized for \(K\) consecutive steps toward
\begin{equation}
    \min_{\delta}\;
    \mathcal{L}_{\mathrm{harm}}
    \bigl(\theta_t+\delta\bigr),
    \label{eq:harmful-estimation}
\end{equation}
where \(\mathcal{L}_{\mathrm{harm}}\) denotes the cross-entropy loss on \(\mathcal{D}_{\mathrm{harm}}\). The resulting perturbation is denoted by \(\Delta h\).

We then hold \(\Delta h\) fixed and perform \(K\) safety alignment steps. At each step \(i\in\{t,\ldots,t+K-1\}\), we independently sample a new retention probability and mask:
\begin{equation}
    p_i\sim\mathcal{U}(0,1),
    \qquad
    M_i\sim\operatorname{Bernoulli}(p_i),
    \qquad
    \Delta h_i=M_i\odot\Delta h.
    \label{eq:jwp-mask}
\end{equation}
For an alignment minibatch \(\mathcal{B}_i\), we minimize
\(\mathcal{L}_{\mathrm{align}}(\theta_i+\Delta h_i;\mathcal{B}_i)\)
to update the alignment parameters while keeping the perturbation fixed. The gradients computed in the perturbed state update \(\theta_i\) to \(\theta_{i+1}\), encouraging the model to maintain safe responses under harmful perturbations.

The same \(\Delta h\) is used for \(K\) alignment steps, with a new mask sampled at each step. We then estimate a fresh perturbation from \(\theta_{t+K}\) for the next cycle. Each cycle therefore uses \(K\) steps of simulated harmful fine-tuning and \(K\) steps of safety alignment, as summarized in Algorithm~\ref{alg:jwp}.

\begin{algorithm}[t]
\caption{Joint Weight Perturbations}
\label{alg:jwp}
\begin{algorithmic}[1]
\Require Initial weights \(\theta_0\), datasets \(\mathcal{D}_{\mathrm{align}}\) and \(\mathcal{D}_{\mathrm{harm}}\), alignment steps \(T\), and steps per phase \(K\)
\Ensure Aligned model weights \(\theta_T\)

\For{\(i=0,\ldots,T-1\)}
    \If{\(i\bmod K=0\)}
        \State Freeze \(\theta_i\)
        \State Initialize fresh temporary parameters with effective perturbation \(\delta=0\)
        \State Fit \(\Delta h\) for \(K\) steps using Equation~\eqref{eq:harmful-estimation}
    \EndIf

    \State Freeze \(\Delta h\) and enable gradients for the alignment parameters
    \State Sample an alignment minibatch \(\mathcal{B}_i\)
    \State Sample \(p_i\sim\mathcal{U}(0,1)\) and \(M_i\sim\operatorname{Bernoulli}(p_i)\)
    \State \(\Delta h_i\gets M_i\odot\Delta h\)
    \State \(\mathcal{L}_i\gets
    \mathcal{L}_{\mathrm{align}}(\theta_i+\Delta h_i;\mathcal{B}_i)\)
    \State Compute gradients of \(\mathcal{L}_i\) with respect to the trainable alignment parameters
    \State Use these gradients to update \(\theta_i\) to \(\theta_{i+1}\)
\EndFor

\State \Return \(\theta_T\)
\end{algorithmic}
\end{algorithm}

\section{Experiments}
\label{sec:experiments}

\subsection{Experiment Setup}
\label{sec:experiment-setup}

\paragraph{Datasets.}
We follow prior work~\citep{huang2025booster} to construct the datasets for safety alignment and downstream fine-tuning. For alignment, we use 5,000 examples curated from BeaverTails~\citep{ji2023beavertails}, paired with safe target responses. Harmful prompt--response pairs from the same source are used by methods requiring auxiliary harmful data. We consider four downstream datasets: SST-2~\citep{wang2018glue}, AGNEWS~\citep{zhang2015character}, GSM8K~\citep{cobbe2021training}, and ARC-Challenge~\citep{clark2018thinksolvedquestionanswering}. To simulate harmful fine-tuning, we mix harmful examples with benign task examples at a default harmful-data ratio of \(10\%\).

\paragraph{Models and baseline methods.}
We conduct experiments on Llama-2-7B~\citep{touvron2023llama}, Llama-3.1-8B~\citep{grattafiori2024llama}, and Qwen2.5-7B~\citep{hui2024qwen2}, starting from their base checkpoints. Following prior works \citep{huang2025booster, nguyen2026antibody}, we use Llama-2-7B and SST-2 as the default model and downstream task. We compare JWP with four alignment-stage baselines: supervised fine-tuning (SFT), Vaccine~\citep{huang2024vaccine}, RepNoise~\citep{rosati2024representation}, and Booster~\citep{huang2025booster}. Baseline details are provided in Appendix~\ref{app:baselines}.

\paragraph{Metrics and evaluation.}
We evaluate safety and utility using three metrics. Harmful Score (HS) measures the percentage of responses classified as harmful by Beaver-Dam-7B on BeaverTails evaluation prompts after harmful fine-tuning. Fine-tuning Accuracy (FA) measures accuracy on the corresponding downstream task. Utility Preservation (UP) measures task accuracy immediately after safety alignment and before downstream fine-tuning, using 5-shot evaluation by default. Comparing UP with the corresponding base model's accuracy allows us to assess the alignment tax \citep{lin2024mitigating} incurred during safety alignment. Since our work targets scenarios with strong downstream attacks, we place emphasis on post-attack HS, while also taking UP and FA into account.

\paragraph{Training details.}
Following prior work~\citep{huang2024vaccine,huang2025booster}, we use LoRA for both safety alignment and downstream fine-tuning, with rank 32 and \(\alpha=4\). Prolonged safety alignment can incur a substantial alignment tax, limiting the alignment budget feasible in practical deployments. We therefore use four epochs of safety alignment for all methods, followed by 30 epochs of downstream harmful fine-tuning with a freshly initialized LoRA adapter. This setting examines whether defenses can withstand sustained harmful fine-tuning when the alignment budget is constrained by the need to preserve model utility. We use AdamW with weight decay \(0.1\) and an effective batch size of 10. The learning rates are \(5\times10^{-4}\) for alignment and \(10^{-5}\) for downstream fine-tuning. All methods use the same alignment data and receive the same downstream examples within each task. For JWP, we set \(K=100\) for both harmful perturbation estimation and the subsequent block of safety alignment.  Additional details on dataset construction, training configurations, and evaluation protocols are provided in Appendix~\ref{app:experimental-details}.

\subsection{Safety across Downstream Attacks}

\paragraph{Suppression of joint amplification.}
We first examine whether JWP reduces the joint amplification observed in Section~\ref{sec:joint-effect}. Figure~\ref{fig:joint-effect} (left) shows that JWP exhibits substantially smaller \(I(q)\) values than existing defenses when larger disjoint subsets are combined, indicating less additional harmfulness from their joint application. Figure~\ref{fig:joint-effect} (right) provides complementary evidence: existing defenses exhibit a sharp increase in HS as more of the harmful perturbation is applied, whereas JWP shows a more gradual increase and maintains a substantially lower HS under the full perturbation. These results suggest that reducing joint amplification helps JWP limit safety degradation as harmful perturbation components accumulate, providing a possible explanation for its stronger resistance to sustained harmful fine-tuning in Figure~\ref{fig:baseline-trajectory}.

% \paragraph{Attack duration.}
% Figure~\ref{fig:baseline-trajectory} traces HS during one continuous Llama-2-7B attack with the SST-2 mixture. The plotted defenses start at about 2--5\% HS. SFT and RepNoise reach roughly 38\% and 44\% by step 1,500, while Vaccine and Booster reach about 50\% and 47\% by step 2,500. JWP remains at or below 3\% through step 2,000 and reaches 11\% at step 3,000; the four comparators finish at roughly 52--62\%. The advantage persists along the attack rather than appearing only at the released checkpoint.

\paragraph{Effect of the harmful-data ratio.}
Table~\ref{tab:harmful-ratio} compares performance across harmful-data ratios from \(0\%\) to \(20\%\). JWP achieves the lowest HS throughout this range, with an average of \(18.82\%\) compared with \(47.24\%\) for Booster, the strongest baseline on average. At the highest ratio, repeated training on 200 harmful examples over 30 epochs produces 6,000 harmful-example exposures. Downstream fine-tuning also involves more parameter updates than safety alignment. Although this sustained harmful fine-tuning weakens all evaluated defenses, JWP continues to achieve substantially lower HS than the strongest baseline at this ratio. Its advantage therefore persists under stronger harmful fine-tuning that substantially degrades existing defenses. JWP also achieves the highest average FA of \(94.61\%\), preserving competitive downstream performance alongside its relative safety advantage.

\begin{center}
\begin{minipage}{\textwidth}
\captionof{table}{Performance under different harmful ratios $r$. Clean means no harmful samples and $r$=0.}
\label{tab:harmful-ratio}
\centering
\footnotesize
\setlength{\tabcolsep}{1.5pt}
\setlength{\arrayrulewidth}{0.35pt}
\renewcommand{\arraystretch}{1.00}
\resizebox{\textwidth}{!}{%
\begin{tabular}{c|cccccc|cccccc}
\toprule
\multirow{2}{*}{Methods} & \multicolumn{6}{c}{Harmful Score} & \multicolumn{6}{c}{Fine-tuning Accuracy} \\
\cmidrule(lr){2-7}\cmidrule(lr){8-13}
 & clean & $r$=0.05 & $r$=0.10 & $r$=0.15 & $r$=0.20 & Average & clean & $r$=0.05 & $r$=0.10 & $r$=0.15 & $r$=0.20 & Average \\
\midrule
SFT      & 7.30 & 57.80 & 68.40 & 71.40 & 74.00 & 55.78 & 94.27 & 94.72 & 94.61 & 94.04 & \textbf{94.61} & 94.45 \\
Vaccine  & 3.00 & 35.20 & 60.10 & 70.70 & 75.30 & 48.86 & \textbf{94.95} & \textbf{95.18} & 94.27 & 93.58 & 94.15 & 94.43 \\
RepNoise & 7.80 & 53.50 & 62.60 & 67.70 & 71.50 & 52.62 & 94.72 & 94.61 & 94.27 & 93.46 & 93.69 & 94.15 \\
Booster  & 6.00 & 22.70 & 62.40 & 71.10 & 74.00 & 47.24 & 94.61 & 94.72 & 94.50 & 94.15 & 94.38 & 94.47 \\
\rowcolor{black!12}
JWP      & \textbf{1.40} & \textbf{4.50} & \textbf{14.70} & \textbf{28.00} & \textbf{45.50} & \textbf{18.82} & 94.61 & 94.72 & \textbf{95.07} & \textbf{94.38} & 94.27 & \textbf{94.61} \\
\bottomrule
\end{tabular}%
}
\end{minipage}
\end{center}

\paragraph{Transfer across downstream tasks.}
Table~\ref{tab:task-transfer} compares performance on SST-2, AGNEWS, GSM8K, and ARC-Challenge. JWP achieves substantially lower HS than all baselines across the four tasks, with an average of \(11.50\%\). Downstream accuracy remains competitive across the evaluated tasks and is highest on SST-2. JWP also achieves the highest UP on SST-2 and ARC-Challenge. Its average UP of \(59.39\%\) is the highest among the compared methods. These results demonstrate stronger resistance across downstream tasks while maintaining competitive utility both before and after downstream fine-tuning.

\begin{center}
\begin{minipage}{\textwidth}
\captionof{table}{Performance analysis for different fine-tuning datasets.}
\label{tab:task-transfer}
\centering
\footnotesize
\setlength{\tabcolsep}{3.0pt}
\setlength{\arrayrulewidth}{0.35pt}
\renewcommand{\arraystretch}{1.05}
\resizebox{\textwidth}{!}{%
\begin{tabular}{l|rrr|rrr|rrr|rrr|rrr}
\toprule
\multirow{2}{*}{Methods}
& \multicolumn{3}{c|}{SST-2}
& \multicolumn{3}{c|}{AGNEWS}
& \multicolumn{3}{c|}{GSM8K}
& \multicolumn{3}{c|}{ARC-Challenge}
& \multicolumn{3}{c}{Average} \\
\cmidrule(lr){2-4}
\cmidrule(lr){5-7}
\cmidrule(lr){8-10}
\cmidrule(lr){11-13}
\cmidrule(lr){14-16}
& HS $\downarrow$ & FA $\uparrow$ & UP $\uparrow$
& HS $\downarrow$ & FA $\uparrow$ & UP $\uparrow$
& HS $\downarrow$ & FA $\uparrow$ & UP $\uparrow$
& HS $\downarrow$ & FA $\uparrow$ & UP $\uparrow$
& HS $\downarrow$ & FA $\uparrow$ & UP $\uparrow$ \\
\midrule
SFT
& 68.4 & 94.61 & 91.17
& 67.4 & \textbf{90.00} & 76.28
& 48.4 & 18.70 & 9.63
& 65.0 & 53.50 & 53.07
& 62.30 & 64.20 & 57.54 \\
Vaccine
& 60.1 & 94.27 & 53.67
& 65.4 & 87.80 & 77.84
& 14.2 & 18.20 & 4.02
& 59.6 & 51.88 & 44.28
& 49.83 & 63.04 & 44.95 \\
RepNoise
& 62.6 & 94.27 & 74.54
& 63.7 & 88.60 & 66.68
& 54.8 & 18.60 & 9.17
& 60.3 & 53.50 & 52.13
& 60.35 & 63.74 & 50.63 \\
Booster
& 62.4 & 94.50 & 87.73
& 63.6 & 89.90 & \textbf{79.87}
& 22.2 & \textbf{20.70} & \textbf{12.66}
& 62.9 & \textbf{54.61} & 52.90
& 52.78 & \textbf{64.93} & 58.29 \\
\rowcolor{black!12}
JWP
& \textbf{14.7} & \textbf{95.07} & \textbf{92.20}
& \textbf{16.5} & 88.90 & 78.84
& \textbf{3.6} & 20.20 & 12.43
& \textbf{11.2} & 53.84 & \textbf{54.10}
& \textbf{11.50} & 64.50 & \textbf{59.39} \\
\bottomrule
\end{tabular}%
}
\end{minipage}
\end{center}

\paragraph{Transfer across model families.}
Table~\ref{tab:model-transfer} compares performance on Llama-2-7B, Llama-3.1-8B, and Qwen2.5-7B under harmful fine-tuning with the ARC-Challenge mixture. JWP achieves substantially lower HS than all baselines on every model. Its average HS is \(11.90\%\), compared with baseline averages ranging from \(49.30\%\) to \(64.93\%\). These safety improvements are accompanied by competitive downstream accuracy across models, with JWP achieving the highest average FA of \(56.68\%\). JWP also achieves the highest UP on Llama-2-7B and Llama-3.1-8B. Its average UP of \(57.57\%\) is close to the highest baseline average of \(57.65\%\), indicating that capability preservation during safety alignment remains competitive.

\begin{center}
\begin{minipage}{\textwidth}
\captionof{table}{Performance across model families under harmful fine-tuning with the ARC-Challenge mixture.}
\label{tab:model-transfer}
\centering
\footnotesize
\setlength{\tabcolsep}{3.0pt}
\setlength{\arrayrulewidth}{0.35pt}
\renewcommand{\arraystretch}{1.05}
\begin{tabular}{l|rrr|rrr|rrr|rrr}
\toprule
\multirow{2}{*}{Methods}
& \multicolumn{3}{c|}{Llama-2-7B}
& \multicolumn{3}{c|}{Llama-3.1-8B}
& \multicolumn{3}{c|}{Qwen2.5-7B}
& \multicolumn{3}{c}{Average} \\
\cmidrule(lr){2-4}
\cmidrule(lr){5-7}
\cmidrule(lr){8-10}
\cmidrule(lr){11-13}
& HS $\downarrow$ & FA $\uparrow$ & UP $\uparrow$
& HS $\downarrow$ & FA $\uparrow$ & UP $\uparrow$
& HS $\downarrow$ & FA $\uparrow$ & UP $\uparrow$
& HS $\downarrow$ & FA $\uparrow$ & UP $\uparrow$ \\
\midrule
SFT
& 65.00 & 53.50 & 53.07
& 67.50 & 58.70 & 55.89
& 53.50 & 55.89 & 59.73
& 62.00 & 56.03 & 56.23 \\
Vaccine
& 59.60 & 51.88 & 44.28
& 48.00 & 56.74 & 39.25
& 58.00 & 58.36 & 58.36
& 55.20 & 55.66 & 47.30 \\
RepNoise
& 60.30 & 53.50 & 52.13
& 68.00 & 57.42 & 56.40
& 66.50 & \textbf{58.53} & 57.94
& 64.93 & 56.48 & 55.49 \\
Booster
& 62.90 & \textbf{54.61} & 52.90
& 55.00 & \textbf{59.04} & 58.28
& 30.00 & 56.06 & \textbf{61.77}
& 49.30 & 56.57 & \textbf{57.65} \\
\rowcolor{black!12}
JWP
& \textbf{11.20} & 53.84 & \textbf{54.10}
& \textbf{11.00} & 58.19 & \textbf{59.04}
& \textbf{13.50} & 58.02 & 59.56
& \textbf{11.90} & \textbf{56.68} & 57.57 \\
\bottomrule
\end{tabular}
\end{minipage}
\end{center}

\subsection{Mechanism Ablations}

\paragraph{Sampling the retention probability.}
At each alignment step, JWP samples \(p\sim\mathcal{U}(0,1)\) and then generates a Bernoulli mask with retention probability \(p\). A natural alternative is to fix \(p\) while still resampling the mask at every step. Table~\ref{tab:occupation-ablation} compares these choices with \(K=100\). Among the fixed settings, HS decreases as \(p\) increases. However, always applying the full perturbation (\(p=1\)) yields an HS of \(31.40\%\), substantially higher than the \(14.70\%\) achieved by resampling \(p\).

This advantage is consistent with the broader coverage provided by randomizing \(p\). A fixed intermediate probability generates different component combinations but concentrates them around a similar retained fraction. For example, \(p=0.5\) makes fully applied perturbations highly unlikely, whereas \(p=1\) excludes all partially applied combinations. Resampling \(p\) instead exposes alignment to combinations spanning different perturbation sizes. These results support the design of JWP: training across diverse harmful perturbation states to improve resistance to harmful fine-tuning, with the aim of suppressing joint amplification among their components. Appendix~\ref{app:mask-analysis} provides a mathematical characterization of the coverage induced by this sampling scheme.

\begin{center}
\begin{minipage}{\textwidth}
\captionof{table}{Effect of the perturbation retention probability during alignment.}
\label{tab:occupation-ablation}
\centering
\small
\setlength{\tabcolsep}{8pt}
\setlength{\arrayrulewidth}{0.35pt}
\renewcommand{\arraystretch}{1.10}
\begin{tabular}{l|cccccc|c}
\toprule
Retention probability \(p\)
& \(0\) & \(0.2\) & \(0.4\) & \(0.6\) & \(0.8\) & \(1\)
& \cellcolor{black!12}\(\mathcal{U}(0,1)\) \\
\midrule
Harmful Score $\downarrow$
& 69.20 & 66.00 & 61.00 & 56.80 & 51.00 & 31.40
& \cellcolor{black!12}\textbf{14.70} \\
\bottomrule
\end{tabular}
\end{minipage}
\end{center}

\paragraph{Length of alternating training blocks.}
To control alignment overhead, JWP uses \(K\) steps to estimate a harmful perturbation and reuses it for the following \(K\) alignment steps. This keeps the total numbers of perturbation-fitting and alignment steps comparable. The fitting cost is thus distributed across multiple alignment steps, whereas TAR performs multi-step fine-tuning simulations for each outer update~\citep{tamirisa2025tamper}.

Table~\ref{tab:field-scale-ablation} examines this choice of \(K\). Short blocks yield high HS, suggesting that too few fitting steps may inadequately simulate the effects of harmful fine-tuning. Increasing \(K\) allows more extensive perturbation fitting but also reduces the refresh frequency. The lowest HS occurs at \(K=250\), while increasing \(K\) to 500 weakens protection. This pattern is consistent with long reuse intervals leaving the perturbation less representative of the evolving model. Utility preservation also varies with \(K\). Across the evaluated settings, UP remains between \(89.91\%\) and \(92.20\%\), showing comparatively modest variation alongside the substantial changes in HS. These results highlight the importance of balancing perturbation fitting and refresh frequency when configuring \(K\) for a given model.

\begin{center}
\begin{minipage}{\textwidth}
\captionof{table}{Effect of the alternating block length on safety and utility preservation.}
\label{tab:field-scale-ablation}
\centering
\small
\setlength{\tabcolsep}{8pt}
\setlength{\arrayrulewidth}{0.35pt}
\renewcommand{\arraystretch}{1.10}
\begin{tabular}{l|cccccc}
\toprule
Block length \(K\)
& 10 & 20 & 50
& \cellcolor{black!12}100 (default)
& 250 & 500 \\
\midrule
Harmful Score $\downarrow$ 
& 68.60 & 65.70 & 54.90
& \cellcolor{black!12}14.70
& \textbf{10.30} & 28.50 \\
Utility Preservation $\uparrow$
& 90.02 & 89.91 & 91.40
& \cellcolor{black!12}\textbf{92.20}
& 90.71 & 90.94 \\
\bottomrule
\end{tabular}
\end{minipage}
\end{center}

\subsection{Efficiency}

Table~\ref{tab:cost} reports training time and peak GPU memory on Llama-3.1-8B using a single A800-80GB GPU. JWP incurs additional alignment cost relative to SFT because it estimates harmful weight perturbations during training. However, its alignment time of \(0.67\) hours is lower than the \(0.84\) hours required by Booster and \(1.65\) hours required by RepNoise. Its peak alignment memory of \(46.67\) GiB is also lower than both methods. Downstream fine-tuning costs remain comparable across all methods in both runtime and memory usage. Since JWP requires no changes to downstream fine-tuning, its additional computation is incurred once during alignment and does not recur for each subsequent customization request.

\begin{center}
\begin{minipage}{0.98\textwidth}
\captionof{table}{System efficiency on Llama-3.1-8B using a single A800-80GB GPU. Memory denotes peak device usage.}
\label{tab:cost}
\centering
\small
\setlength{\tabcolsep}{5pt}
\setlength{\arrayrulewidth}{0.35pt}
\renewcommand{\arraystretch}{1.05}
\begin{tabular}{l|rrr|rrr}
\toprule
\multirow{2}{*}{Methods}
& \multicolumn{3}{c|}{Wall-clock Time (h)}
& \multicolumn{3}{c}{GPU Memory (GiB)} \\
\cmidrule(lr){2-4}
\cmidrule(lr){5-7}
& Alignment & Fine-tuning & Total
& Alignment & Fine-tuning & Max \\
\midrule
SFT
& 0.25 & 0.27 & 0.52
& 44.10 & 41.94 & 44.10 \\
Vaccine
& 0.49 & 0.27 & 0.76
& 44.97 & 41.94 & 44.97 \\
RepNoise
& 1.65 & 0.28 & 1.93
& 52.28 & 41.94 & 52.28 \\
Booster
& 0.84 & 0.28 & 1.12
& 47.09 & 41.94 & 47.09 \\
\rowcolor{black!12}
JWP
& 0.67 & 0.28 & 0.95
& 46.67 & 41.94 & 46.67 \\
\bottomrule
\end{tabular}
\end{minipage}
\end{center}

\section{Discussion and Limitations}
\label{sec:discussion}

\paragraph{Connection to safety basins.}
Figure~\ref{fig:joint-effect} (right) shows a rapid increase in HS as more components of a harmful perturbation are activated. This resembles the safety-basin phenomenon, where safety drops sharply as random weight perturbations increase~\citep{peng2024navigating}. Although the two settings vary different aspects of perturbation, their similarity suggests that joint amplification may reflect a broader property of safety-relevant parameter space.

\paragraph{Quantifying joint amplification.}
Our analysis measures joint amplification through HS, which captures its behavioral consequence rather than an intrinsic property of the model. Developing model-level measures of these interactions and explaining why they arise remain important directions for future work.

\paragraph{Limitations.}
The choice of \(K\) requires further exploration across models and training settings. Our evaluation focuses on harmful fine-tuning, leaving related threats such as jailbreaks unexplored. We also evaluate JWP independently, while its combination with defenses applied at other stages may provide complementary protection.

\section{Conclusion}

Harmful fine-tuning can undermine safety through the joint amplification of weight perturbations: components with limited individual effects can cause substantially greater harm when activated together. Motivated by this observation, JWP estimates harmful perturbations from the evolving model and trains it to remain safe across masked combinations of their components. Experiments across attack trajectories, data mixtures, downstream tasks, and model families show stronger resistance to harmful fine-tuning while largely preserving utility. These results identify joint amplification as an important target for safety alignment.

\subsection*{AI use statement}
Generative AI tools were used to assist with manuscript drafting and language refinement, including improving clarity, grammar, and readability. The authors take full responsibility for the accuracy and integrity of the final manuscript.

\bibliography{iclr2027_conference}
\bibliographystyle{iclr2027_conference}

\appendix

\section{Experimental Details}
\label{app:experimental-details}

\subsection{Datasets}
\label{app:datasets}

\paragraph{Safety alignment data.}
Following prior work~\citep{huang2024vaccine,huang2025booster}, we construct the alignment dataset from BeaverTails~\citep{ji2023beavertails}. The dataset contains 5,000 examples with safety annotations and paired responses. For benign examples, the alignment target is the original response. For harmful examples, the target is the paired safe response. Methods requiring auxiliary harmful data use the original unsafe responses associated with harmful examples. In JWP, these responses are used to estimate harmful weight perturbations, while the corresponding safe targets are used for safety alignment.

\paragraph{Downstream fine-tuning data.}
We consider SST-2~\citep{wang2018glue}, AGNEWS~\citep{zhang2015character}, GSM8K~\citep{cobbe2021training}, and ARC-Challenge~\citep{clark2018thinksolvedquestionanswering}. These datasets cover binary sentiment classification, four-way news classification, mathematical reasoning, and multiple-choice science question answering, respectively. We report length-normalized accuracy on ARC-Challenge using the \texttt{acc\_norm} metric. For each task, we construct a downstream fine-tuning dataset by mixing benign task examples with harmful examples from BeaverTails. The default dataset contains 1,000 examples with a harmful-data ratio of \(10\%\), consisting of 900 benign task examples and 100 harmful examples. All methods receive the same ordered downstream examples within each task. 

\subsection{Models and Training Configuration}
\label{app:training-details}

We evaluate the base checkpoints of Llama-2-7B, Llama-3.1-8B, and Qwen2.5-7B~\citep{touvron2023llama,grattafiori2024llama,hui2024qwen2}. Unless otherwise specified, Llama-2-7B and SST-2 are the default model and downstream task.

All methods undergo four epochs of safety alignment, followed by 30 epochs of downstream harmful fine-tuning. Prolonged safety alignment can incur a substantial alignment tax, limiting the alignment budget feasible in practical deployments. This configuration evaluates whether defenses remain effective under sustained harmful fine-tuning when safety alignment must also preserve model utility.

We use LoRA for both stages, applied to the query, key, and value projections. Each aligned model receives a freshly initialized LoRA adapter for downstream fine-tuning. The shared configuration is summarized in Table~\ref{tab:app-training-config}. Method-specific objectives follow the corresponding implementations.

\begin{table}[t]
    \centering
    \caption{Shared configuration for safety alignment and downstream fine-tuning.}
    \label{tab:app-training-config}
    \begin{tabular}{lcc}
        \toprule
        Configuration & Safety alignment & Downstream fine-tuning \\
        \midrule
        Training epochs      & 4 & 30 \\
        Optimizer            & AdamW & AdamW \\
        Learning rate        & \(5\times10^{-4}\) & \(10^{-5}\) \\
        Effective batch size & 10 & 10 \\
        Weight decay         & \(0.1\) & \(0.1\) \\
        LoRA rank            & 32 & 32 \\
        LoRA \(\alpha\)      & 4 & 4 \\
        LoRA dropout         & 0 & 0 \\
        Target projections   & Query, key, value & Query, key, value \\
        \bottomrule
    \end{tabular}
\end{table}

\paragraph{JWP configuration.}
For JWP, we set \(K=100\) for both harmful perturbation estimation and the subsequent block of safety alignment. The temporary perturbation is fitted with a learning rate of \(10^{-4}\), using a separate LoRA parameterization with the same rank and scaling as the alignment adapter.

During perturbation estimation, the current alignment parameters remain frozen. During safety alignment, the fitted perturbation remains fixed and only the alignment parameters are optimized. A new retention probability and mask are sampled at every alignment iteration.

\subsection{Metrics and Evaluation}
\label{app:evaluation}

\paragraph{Harmful Score.}
Harmful Score (HS) is the percentage of generated responses classified as harmful by Beaver-Dam-7B on BeaverTails evaluation prompts. We use a classification threshold of \(0.5\). Responses are generated with greedy decoding and a maximum of 512 new tokens. Unless otherwise specified, HS is measured after downstream harmful fine-tuning. For trajectory evaluations, the same metric is also measured at intermediate checkpoints.
For all primary HS results, we use the prompts from the first 1,000 unsafe-labelled examples in the BeaverTails \texttt{30k\_test} split, preserving their dataset order and using the identical  fixed set across all methods; generated responses are scored by the released Beaver-Dam-7B QA-moderation model \citep{ji2023beavertails}.

\paragraph{Fine-tuning Accuracy.}
Fine-tuning Accuracy (FA) measures performance on the corresponding downstream task after fine-tuning. SST-2 and AGNEWS are evaluated by classification accuracy over their respective label sets. GSM8K is evaluated by whether the final numerical answer matches the reference answer. ARC-Challenge is evaluated using \texttt{acc\_norm}, which measures accuracy by selecting the answer option with the highest length-normalized log-likelihood.

\paragraph{Utility Preservation.}
Utility Preservation (UP) measures task accuracy immediately after safety alignment and before downstream fine-tuning. We use 5-shot evaluation by default. UP is reported as accuracy, rather than a ratio normalized by base-model performance. Comparing it with the corresponding base checkpoint's accuracy quantifies the capability loss introduced by safety alignment. This separates the alignment tax from the model's ability to learn a downstream task, which is measured by FA.

\section{Baseline Methods}
\label{app:baselines}

We compare JWP with SFT, Vaccine, RepNoise, and Booster. These methods are applied during safety alignment, after which all models undergo the same downstream fine-tuning procedure. This comparison evaluates the protection learned during alignment under matched downstream training conditions.

\paragraph{Supervised Fine-Tuning (SFT).}
SFT directly optimizes the model on the alignment dataset using the cross-entropy loss. It learns the provided safe targets without introducing simulated perturbations or additional defense objectives. The resulting model subsequently undergoes downstream fine-tuning on the mixture of benign and harmful examples.

\paragraph{Vaccine.}
Vaccine~\citep{huang2024vaccine} introduces perturbations to hidden representations during safety alignment. It constructs perturbations that increase the alignment loss and trains the model to maintain the alignment targets under these perturbations. Its design aims to reduce the sensitivity of aligned behavior to representation changes induced by subsequent harmful fine-tuning.

\paragraph{RepNoise.}
RepNoise~\citep{rosati2024representation} uses harmful examples to suppress harmful information in the model's internal representations. It combines learning from safe targets with objectives that discourage harmful responses and move harmful representations toward a Gaussian noise distribution. The resulting alignment aims to make harmful behavior more difficult to recover through subsequent fine-tuning.

\paragraph{Booster.}
Booster~\citep{huang2025booster} simulates a harmful weight perturbation using gradients computed on harmful examples. In addition to the alignment loss, it penalizes the reduction in harmful loss induced by this perturbation. This objective aims to limit how effectively subsequent harmful fine-tuning can optimize the model toward unsafe responses.

\section{Comparison with TAR}
\label{app:tar-comparison}

TAR~\citep{tamirisa2025tamper} develops resistance to weight modification by training against simulated fine-tuning trajectories. In its safety-response setting, it evaluates a DPO loss at sampled states along these trajectories and combines this signal with a capability-preservation objective. Its knowledge-restriction setting instead uses an entropy-based objective. DPO is therefore one task-specific instantiation of the broader TAR framework.

JWP focuses on the joint effects of harmful perturbation components. Our analyses examine how different parts of the same harmful perturbation affect safety when applied separately and together. They reveal that combinations of components can cause substantially greater degradation than their individual effects suggest. JWP incorporates this observation into alignment by randomly varying both the composition and the applied fraction of a fitted harmful perturbation.

The distinction concerns how the training states are constructed. TAR samples states reached along simulated fine-tuning trajectories. JWP constructs states of the form
\[
    \theta + M\odot\Delta h,
\]
which selectively combine components of the same fitted perturbation. These states need not correspond to checkpoints on the trajectory used to estimate \(\Delta h\). They allow alignment to address combinations that would not be separately evaluated by following that trajectory alone.

The central contribution of JWP is therefore the identification of collective safety degradation and the use of random perturbation coverage to suppress its joint amplification. Simulating harmful fine-tuning provides the perturbation, while random masking determines the combinations against which the model is trained.

\section{Properties of Random Perturbation Coverage}
\label{app:mask-analysis}

This section characterizes the sampling distribution used by JWP. We consider a fixed harmful perturbation \(\Delta h\) partitioned into \(d\) disjoint components, each controlled by one mask variable. Let \(S\subseteq[d]\) denote the retained components. JWP samples \(p\sim\mathcal{U}(0,1)\) and retains each component independently conditional on \(p\).

\subsection{Distribution of Subset Sizes}

For a particular subset \(S\) containing \(k\) components, its sampling probability is
\begin{equation}
    \Pr(S)
    =
    \int_0^1 p^k(1-p)^{d-k}\,dp
    =
    \frac{k!(d-k)!}{(d+1)!}.
    \label{eq:app-subset-probability}
\end{equation}
There are \(\binom{d}{k}\) subsets of size \(k\), so
\begin{equation}
    \Pr(|S|=k)
    =
    \binom{d}{k}\frac{k!(d-k)!}{(d+1)!}
    =
    \frac{1}{d+1},
    \qquad k=0,\ldots,d.
    \label{eq:app-uniform-size}
\end{equation}
Thus, the number of retained components is uniformly distributed. Conditional on this number, all subsets of that size are equally likely.

The JWP objective can consequently be expressed as
\begin{equation}
    \mathcal{L}_{\mathrm{JWP}}(\theta\mid\Delta h)
    =
    \frac{1}{d+1}
    \sum_{k=0}^{d}
    \frac{1}{\binom{d}{k}}
    \sum_{\substack{S\subseteq[d]\\|S|=k}}
    \mathcal{L}_{\mathrm{align}}
    \bigl(\theta+\Delta h_S\bigr).
    \label{eq:app-size-average}
\end{equation}
This expression describes the expected objective. In practice, each alignment iteration uses one sampled mask to estimate it.

A fixed retention probability \(p_0\) produces a binomial distribution over subset sizes, with mean \(dp_0\). Sampling \(p\) uniformly instead assigns equal probability to each possible subset size. This is the role of varying \(p\): alignment is exposed to different extents of perturbation application, including intermediate states.

\subsection{Co-occurrence of Perturbation Components}

For any specified group \(U\) containing \(r\) components, the probability that all its components are retained together is
\begin{equation}
    \Pr(U\subseteq S)
    =
    \mathbb{E}_{p}[p^r]
    =
    \int_0^1p^r\,dp
    =
    \frac{1}{r+1}.
    \label{eq:app-coactivation}
\end{equation}
Other components may also be retained in this event. With a fixed retention probability \(p_0\), the corresponding probability is \(p_0^r\). These expressions describe how the choice of retention distribution changes the frequency with which particular groups appear together during training.

The sampling design is motivated by the observations in Section~\ref{sec:joint-effect}: perturbation subsets with limited individual effects can produce substantial safety degradation when combined. Varying the mask allows alignment to evaluate different combinations, while varying \(p\) changes their extent of application. The identities above characterize this coverage. Whether training with it reduces the observed joint amplification during subsequent harmful fine-tuning is evaluated experimentally.

\end{document}